\pdfoutput=1

\documentclass{llncs}

\usepackage{graphicx}% http://ctan.org/pkg/graphicx
\usepackage{xcolor}
\usepackage{amsfonts}
\usepackage{booktabs} % For formal tables
\usepackage{amsfonts}
\usepackage{amssymb}
\usepackage{xspace}
\usepackage{epstopdf}
\usepackage{enumitem}
\usepackage{algorithm}
\usepackage{algorithmic}
\usepackage{multirow}
\usepackage{booktabs}
\usepackage{array}
\usepackage{wrapfig}
\usepackage{pifont}
\usepackage{pbox}
\usepackage{amsmath}

\begin{document}

\title{An EEG-based Image Annotation System}
\titlerunning{An EEG-based Image Annotation System}  % abbreviated title (for running head)
\author{Viral Parekh\inst{1} \and Ramanathan Subramanian\inst{2} \and 
Dipanjan Roy\inst{3} \and C. V. Jawahar\inst{1}}
\authorrunning{Parekh et al.} % abbreviated author list (for running head)
\tocauthor{Viral Parekh, Ramanathan Subramanian, Dipanjan Roy, and C. V. Jawahar}
\institute{IIIT Hyderabad, India,\\
\and
University of Glasgow, Singapore\\
\and 
National Brain Research Centre, Manesar, India}

\maketitle              % typeset the title of the contribution

\begin{abstract}The success of deep learning in computer vision has greatly increased the need for annotated image datasets. We propose an EEG (Electroencephalogram)-based image annotation system. While humans can recognize objects in 20-200 milliseconds,  the need to manually label images results in a low annotation throughput. Our system employs brain signals captured via a consumer EEG device to achieve an annotation rate of up to 10 images per second.  We exploit the P300 event-related potential (ERP)  signature to identify target images during a rapid serial visual presentation (RSVP) task. We further perform unsupervised outlier removal to achieve an F1-score of 0.88 on the test set. The proposed system does not depend on  category-specific EEG signatures enabling the annotation of any new image category without any model pre-training.
\end{abstract}
\section{Introduction}
Image annotation is a critical task in computer vision, intended to bridge the \textit{\textbf{semantic gap}} between automated and human understanding via the use 
of tags and labels. Image annotation is useful for building large-scale retrieval systems, organizing and managing multimedia databases, and for training deep learning models for scene understanding. A trivial way to annotate images is to tag them manually with the relevant labels, but this approach is slow and tedious for huge databases. Therefore, many efforts have been undertaken to address/circumvent this problem. Some methods are completely automatic \cite{zhang2010automatic,verma2012image,verma2013exploring,pascal,fu2012random}, while some others are interactive \cite{wang2011active,1035755,bakliwal2015active,Katti2010,Subramanian2014}-- these approaches have considerably reduced the human effort required for annotation.

% A typical active learning system few samples most informative unlabeled images are selected for manual annotation and then these images are added to the training set for model learning. Hence this is one way to reduce human effort in image annotation task. \\

Human vision is a very powerful system for object recognition and scene understanding. It is also robust to variations in illumination, scale or pose. We are habitually used to recognizing objects even in cluttered scenes. Humans can identify objects in tens of milliseconds \cite{gistofscene,speedofsight}, but the representation of the perceived information via hand movements or verbal responses for annotation is very slow compared to the processing speed of contemporary digital devices. In this regard, the emerging field of brain-Computer Interfaces (BCI) offers us an innovative way to exploit the power of human brain for data annotation with minimal effort. 

Brain-Computer Interfaces rely on various technologies for sensing brain activity such as Electroencephalography (EEG), MEG (Magnetoencephalography), PET (Positron Emission Tomography), SPECT (Single Photon Emission Computed Tomography), fMRI (functional Magnetic Resonance Imaging) and fNIRS (functional near infrared spectroscopy). Among these, EEG provides a high temporal resolution (sampling rate of up to 1 KHz) and adequate spatial resolution (1-2 cm). In this work, we specifically use the portable and easy-to-use consumer grade \textit{Emotiv} EEG device, which enables a minimally intrusive user experience as users perform cognitive tasks, for sensing and recording brain activity. While having these advantages, consumer EEG devices nevertheless suffer from a high signal-to-noise ratio, which  makes subsequent data analytics challenging.

In this work, we focus on the annotation of a  \textit{pre-selected} object category over the entire image dataset instead of labeling all categories at once. If the images are presented serially in a sequence for annotation, then the task is equivalent to that of \textit{target detection}. Now whenever an image containing a target class instance is observed by the human annotator, an event-related potential (ERP) signature known as \textbf{P300} \cite{linden2005p300} is observed in the EEG data. By examining the EEG signals generated during image presentation, we can discover the images of interest and annotate them accordingly. In this paper, we provide the pipeline and architecture for image annotation via EEG signals.

\section{Related work}
The use of EEG  as an additional modality for computer vision and scene understanding tasks has been explored by a number of works. In \cite{seg}, EEG signals are used to automate grab cut-based image segmentation. In \cite{isearch}, authors exploit ERP signatures such as P300 for image retrieval. {In \cite{koelstra2009eeg}, authors use the N400 ERP to validate tags attached to video content.} Emotions from movies and ads are inferred via EEG signals in~\cite{subramanian2016ascertain} and~\cite{Shukla2017acm}. 

Few studies directly use image category-based EEG signatures for recognizing aspects related to multimedia content as well as users. For example, the authors of \cite{msoft}  use EEG signals to classify images into three object categories-- animals, faces and inanimate. In a recent work \cite{mshah}, the authors present how EEG features can be employed for multi-class image classification. Another recent work recognizes user gender from EEG responses to emotional faces~\cite{Bilalpur17}. Given the state-of-the-art, the key contributions of our work are we how how (i) the P300 ERP signature can be employed for image annotation; (ii) the model trained for one object category can be directly used for a novel category, and (iii) the image presentation time affects annotation system performance for complex images. 

\section{System architecture}

\begin{figure*}[!ht]
\begin{center}
\includegraphics[width=0.8\linewidth]{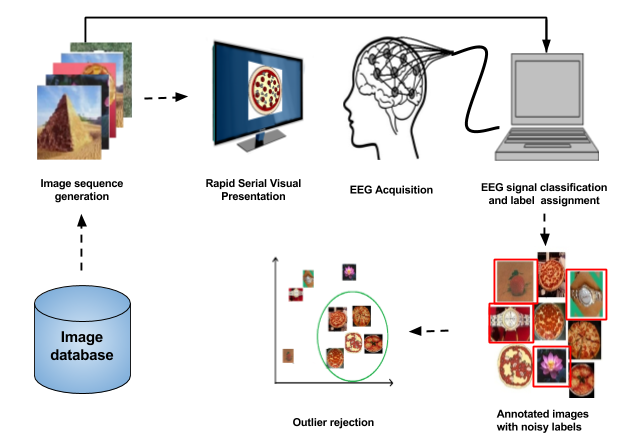}\vspace{-.1cm} 
\end{center}
\caption{\textbf{EEG-based annotation pipeline:} An exemplar illustration for the \textit{pizza} object class is presented. Best viewed in color and under zoom.} 
\label{fig:sysarch}
\vspace{-.3cm}
\end{figure*}

The proposed image annotation system consists of several components-- RSVP generation, EEG data acquisition, EEG pre-processing, (binary) classification and outlier removal. Fig.\ref{fig:sysarch} presents an overview of the EEG-based annotation pipeline. The RSVP generation unit prepares the set of images for viewing, so that a few among those correspond to the target object category. The image sequence is created via random sampling from the whole dataset. A human annotator is then asked to identify the target category images as the sequence is presented rapidly, and the annotator's brain activity is recorded via an EEG headset during the visual recognition task. The compiled EEG data is first pre-processed for artifact removal. Then, the classification unit categorizes the EEG responses into \textit{target} and \textit{non-target} annotations based on P300 patterns. Images classified as \textit{target} are annotated with the target label class. However, this labeling is noisy due to the presence of false positives and imbalance towards the negative (non-target) class. An outlier removal unit finally performs unsupervised dimensionality reduction and clustering to improve the labeling precision.

%----------------------------------bci--------------------------------

\subsection{Rapid Serial Visual Presentation and Oddball paradigm}
\begin{figure}[!ht]
\begin{center}
\fbox{\includegraphics[width=0.6\linewidth]{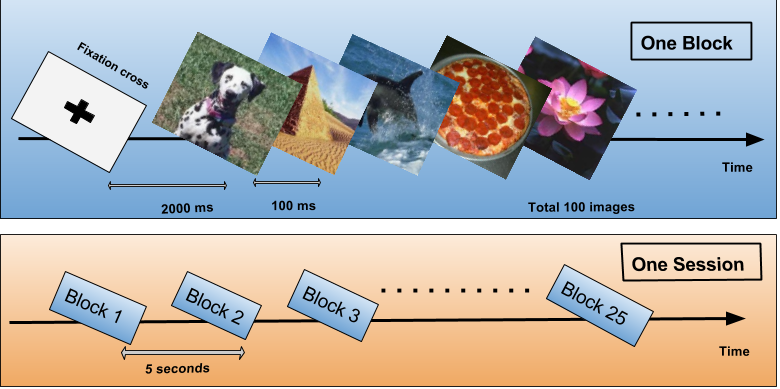}}
\caption{\textbf{Experimental protocol:} Participants completed two identical sessions (one used for training and the other for test) which were 5 minutes apart. Each session comprised 25 blocks of 100 images, and lasted about six minutes.}
\label{fig:rsvp_protocol}
\end{center}
\end{figure}

Rapid Serial Visual Presentation is popularly used in psychophysical studies, and involves a series of images or other stimuli types being presented to viewers with a speed of around 10 items per second. This paradigm is basically used to examine the characteristics pertaining to visual attention. In RSVP studies, the \textit{oddball} phenomenon~\cite{picton1992p300} is widely  used. In the oddball paradigm, a deviant (target) stimulus is infrequently infused into a stream of audio/visual stimuli. For EEG-based annotation, we generated an RSVP sequence by combing a few \textit{target} category images with many \textit{non-target} images via random sampling from the original dataset. Each image in the sequence was then shown to the viewer for 100 ms, and a fixation cross was presented for 2 seconds at the beginning of the sequence to minimize memory effects and to record resting state brain activity (see Fig.\ref{fig:rsvp_protocol}). 

\subsection{EEG data preprocessing and classification}

\begin{figure}[!h]
\begin{center}
\includegraphics[height=4cm]{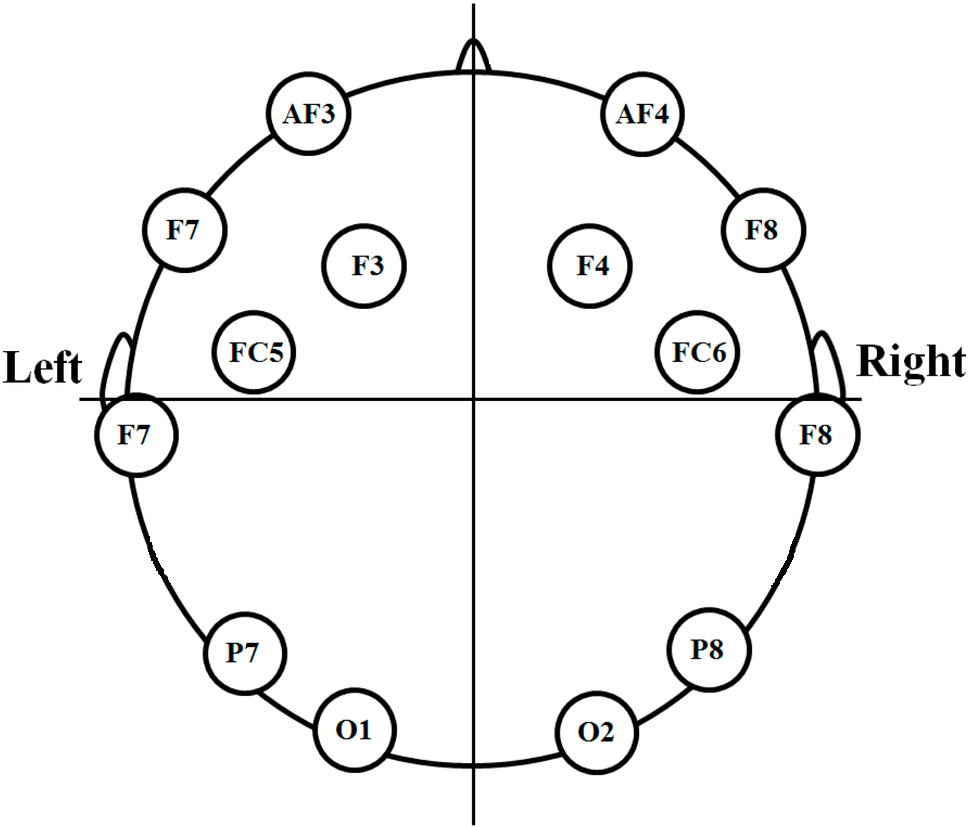}\vspace{-.1cm}
\end{center}
\caption{\textbf{Sensor configuration:} Emotiv electrode locations as per International 10-20 system.}
\label{fig:chanlocs}
\vspace{-.2cm}
\end{figure}

We used the \textit{Emotiv EPOC} headset to record EEG data. This is a 14 channels (plus CMS/DRL references, P3/P4 locations) Au-plated dry electrode system. For ERP analysis, the Emotiv provides signals comparable to superior lab-grade EEG devices with 32, 64 or 128 channels. The headset uses sequential sampling at 2048 Hz internally which is down-sampled to 128 Hz. The incoming signal is automatically notch filtered at 50 and 60 Hz using a $5^{th}$ order sinc notch filter. The resolution of the electrical potential is 1.95 $\mu$V. The locations for the 14 channels are as per International 10-20 locations as shown in Fig.\ref{fig:chanlocs}.
% af3, f7, f3, fc5, t7, p7, o1, o2, p8, t8, fc6, f4, f8 and af4

% \begin{figure}[!tbp]
%   \centering
%   \begin{minipage}[b]{0.4\textwidth}
%     \includegraphics[width=\textwidth]{images/emotiv.png}
%     \caption{\textbf{Sensor configuration:} Emotiv electrode locations as per International 10-20 system.}
% \label{fig:chanlocs}
%   \end{minipage}
%   \hfill
%   \begin{minipage}[b]{0.4\textwidth}
%     \includegraphics[width=\textwidth]{images/erps.png}
%     \caption{\textbf{ERP plots:} ERP curves for the Emotiv af3, af4, f3 and f4 channels for \textit{target} (red) and \textit{not-target} (blue) images. P300 signatures are evident for targets but not for non-targets.} 
% \label{fig:p3erp}
%   \end{minipage}
% \end{figure}

The recorded EEG data is contaminated by various noise undesirable signals that originate from outside the brain. For instance, while recording EEG, one often encounters 50/60Hz power-line noise and artifacts caused by muscle or eye movements. We extracted one second long \textit{epochs} corresponding to each 100 ms long \textit{trial} denoting the presentation of an image, with 128Hz sampling rate. Our EEG preprocessing includes (a) baseline power removal using the 0.5 second pre-stimulus samples, (b) band-pass filtering in 0.1-45Hz frequency range, (c) independent component analysis (ICA) to remove artifacts relating to eye-blinks, and eye and muscle movements. Muscle movement artifacts in EEG are mainly concentrated between 40-100 Hz. While most artifacts are removed upon EEG band-limiting, the remaining are removed manually via inspection of ICA components.

\begin{figure}[!h]
\begin{center}
\includegraphics[width=0.9\linewidth]{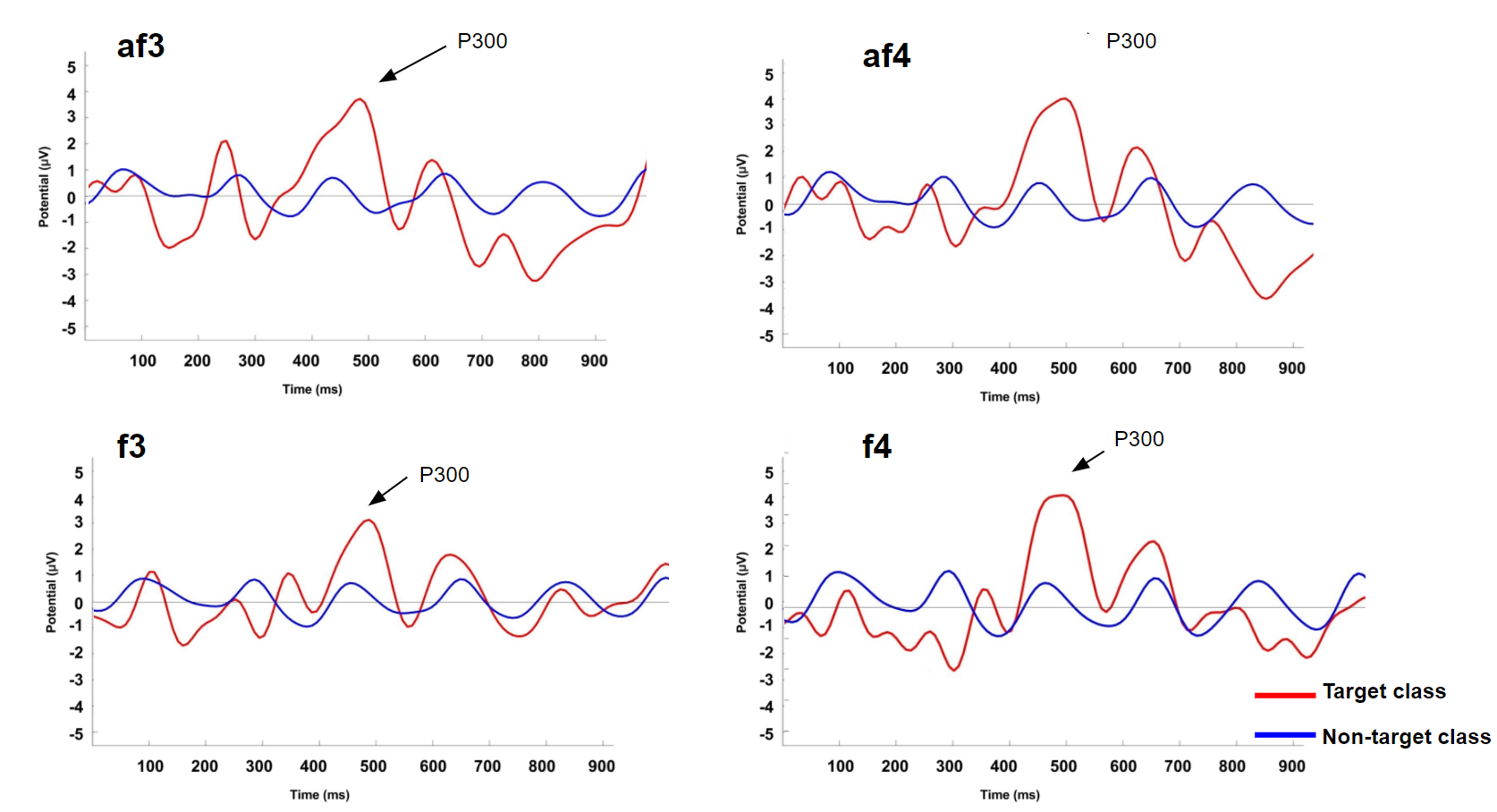}
% * <parekh.viral@research.iiit.ac.in> 2017-11-01T12:12:43.870Z:
%
% ^.
\end{center}
\caption{\textbf{ERP plots:} ERP curves for the Emotiv af3, af4, f3 and f4 channels for \textit{target} (red) and \textit{not-target} (blue) images. P300 signatures are evident for targets but not for non-targets.} 
\label{fig:p3erp}
\vspace{-.1cm}
\end{figure}

The human brain's response to a stimulus can be measured as a voltage fluctuation resulting from the ionic current within the neurons. The event-related potential is one such measure that is directly related to some motor, cognitive or sensory activation. Out of various ERP components, the P300 signature is commonly elicited in the oddball paradigm where very few targets are mixed with a large number of non-targets. In our experimental setup, we employed a 1:12 ratio for target-to-non-target images. As shown in Fig.\ref{fig:p3erp}, the P300 ERP signature is observed between 250 to 500 ms post \textit{target} stimulus presentation. Also, the ERP response is significantly different for target and non-target images, and therefore can be exploited for EEG-based image annotation. 

We used the Convolutional Neural Network (CNN)-based EEGNet architecture \cite{eegnet} to classify our EEG data based on P300 detection in the RSVP task. The EEGnet architecture consists of only three convolutional layers. All layers use the Exponential Linear Unit (ELU) \cite{elu} as nonlinear activation function with parameter $\alpha=1$. We trained the model using the minibatch gradient descent algorithm with categorical cross-entropy criterion and Adam optimizer \cite{kingma2014adam}. The models were trained on a NVIDIA GEFORCE GTX 1080 Ti GPU, with CUDA 8 and cuDNN v6 using the Pytorch \cite{paszkepytorch} based Braindecode \cite{braindecode} library.
% \vspace{5cm}
\subsection{Outlier removal}
{We select one category at a time for the annotation task, which results in class imbalance for the RSVP task. The selected object category forms the \textit{target} class, while all other categories collectively form the \textit{non-target} class.} Due to this heavy class imbalance and the characteristics of P300 as discussed in Section \ref{Res}, the false positive rate of the predicted labels is high. Therefore we performed unsupervised outlier removal on the predicted \textit{target} images. Deep learning features have proven advantages over hand-crafted features like SIFT and HoG  \cite{deepfeatures}. We used a pre-trained VGG-19 model \cite{vgg} to obtain the feature descriptors for the targets. These feature descriptors provide compact representation of raw images while preserving the information required to distinguish between image classes. Each target image was fed forwarded within the VGG-19 model to obtain the 4096 dimensional feature vectors. Target images need not belong to the image classes on which the model is pre-trained. Then, we perform dimensionality reduction with t-SNE \cite{maaten2008visualizing} to generate low-dimensional features. The t-SNE algorithm retains the local structure of the data while also revealing some important global structure, and hence it performs better than principal component analysis (PCA) alone. 

In our case, we assume that samples from the target class should be close in feature space as compared to non-target samples. By performing a grid search on hyper-parameters, we found that the algorithm works best with perplexity value 20, 50 PCA components and 3-5 output dimensions. Then, we performed \textit{k}-means clustering for two classes assuming that target class samples will form a cluster distinct from the false positives. Also, since the false positive cluster would contain samples from many categories, the cluster would not be as dense as the target cluster.

%----------------------------------Results--------------------------------
\section{Protocol design and Experiments}

\subsection{Datasets}
To evaluate the performance of our image annotation system, we used the Caltech101 (CT) \cite{fei2007learning} and Pascal VOC2012 (PV) \cite{pascal-voc-2012} datasets. The CT dataset consists of 101 object categories with 40 to 800 images per category. The PV dataset contains a total of 11,530 images from 20 categories, and multiple object categories can be present in one image. 

\subsection{Experimental setup}
We utilized 2500 images for training, and 2500 images for testing. Both these image sets comprised 200 images of a particular target category that we wanted to annotate. All images were resized 512$\times$512 pixels, and images were displayed at 10 Hz frequency in blocks of 100 in order to minimize viewer distraction and fatigue. During the RSVP task, participants were shown a fixation display for 2 seconds at the beginning of each 100 image sequence. Train and test EEG data were captured using an identical experimental setup with the temporal gap of 5 minutes. Target image categories were decided \textit{a priori} before every experiment. 

Our study was conducted with five graduate students (5 male, age 24.4 $\pm$ 2.1) with 10/20 corrected vision, seated at a distance of 60 cm from the display. A total of three sessions (each involving train and test set) were performed with each participant. To facilitate engagement, viewers were instructed to count the number of target images during the experiment. Target image classes were different for each session, and included categories like \textit{bike}, \textit{pizza}, \textit{panda}, \textit{sofa}, \textit{etc}. Each participant performed two sessions on the CT dataset and one session on the PV dataset.

\section{Results and Discussion}\label{Res}

Due to a heavy class imbalance between \textit{target} and \textit{non-target} category images, we use the F1-score to evaluate our annotation results. The F1-score is a popular performance metric used in retrieval studies, and denotes the harmonic mean of the precision and recall scores.  
%F1 score is calculated by $2 (precision \times recall) / (precision + recall)$. 
All reported results denote the mean F1 achieved with five-fold cross validation.

\begin{table}[!h]
\begin{center}
\renewcommand{\arraystretch}{1.3}
\caption{\textbf{Results synopsis:} Annotation performance obtained for the CT and PV datasets across total 15 sessions (5 viewers).}
\label{table:mresults}
\begin{tabular}{|l|c|c|}
\hline
\textbf{Dataset} & \textbf{Caltech101} & \textbf{Pascal VOC 2012} \\
\hline\hline
\textbf{Before outliers removal} & & \\
F1 score& 0.71 & 0.68\\
Precision& 0.66 & 0.63\\
Recall& 0.81 & 0.72\\
\hline
\textbf{After outliers removal} & & \\
F1 score& \textbf{0.88} & \textbf{0.83}\\
Precision& 0.99 & 0.97\\
Recall& 0.81 & 0.72\\
\hline
Target image percentage & 8\% & 8\%\\
Image presentation speed & 10 Hz & 10 Hz\\
Number of images in test set & 2500 & 2500\\
\hline
\end{tabular}
\end{center}
\vspace{-.2cm}
\end{table}

In Table~\ref{table:mresults}, we report the averaged F1 and precision-recall values for the CT and PV datasets across all participants. Note that the precision and F1 scores improve significantly upon outlier removal due to a stark reduction in the number of false positives via feature-based clustering. Overall F1 scores for the PV dataset are lower than for the CT dataset. This can be attributed to the fact that the PV dataset is more complex, as it contains multiple object classes in many images, as compared to CT which contains only one object class per image.

As our annotation system is dependent on viewer ability, its performance is sensitive to human factors. One key factor is the image presentation rate. The image display latency (100 ms) is lower than the P300 response latency ($\approx$ 300 ms)~\cite{polich2007updating}. The rapid image display protocol results in (i) viewers confusing between similar object classes, (ii) viewers unable to fully comprehend visual information from complex images, and (iii) EEG data for consecutive images having significant overlap leading to misclassification. 

\begin{figure}[!h]
\begin{center}
\includegraphics[width=0.7\linewidth]{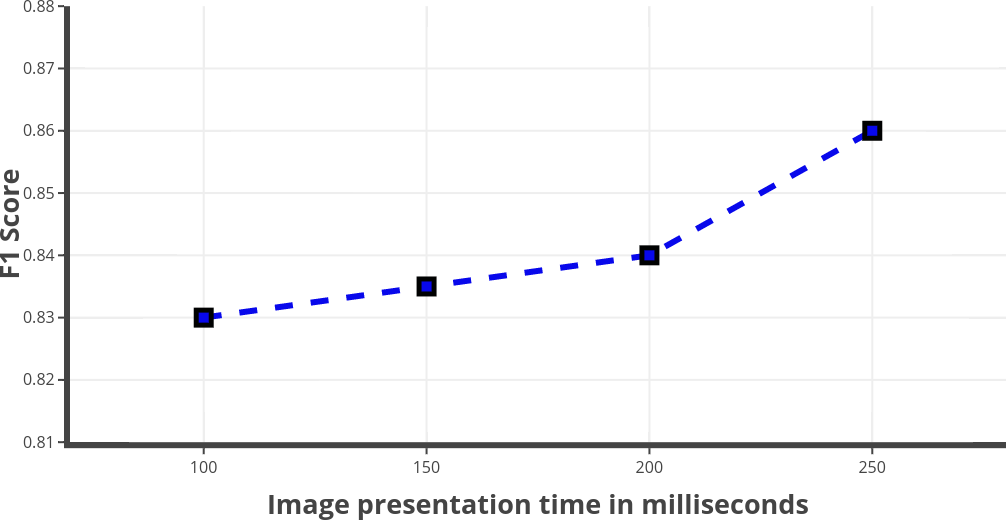}
\end{center}
\caption{\textbf{Presentation rate vs annotation performance:} Variation in F1-score with image display rate.}
\label{fig:ispeed}
\vspace{-0.5cm}
\end{figure}

Therefore, we hypothesized that reducing the image display rate would (a) allow the viewer to better comprehend the visual content (especially for complex images), (b) better delineation of EEG responses, and (c) better manifestation of ERP signatures. These in turn, would improve our annotation performance while marginally reducing the annotation throughput. Fig.\ref{fig:ispeed} presents the observed results. Note that a 3\% increase in F1-score is observed when the image presentation rate is reduced from 10 to 4 images/second, validating our hypothesis.

Conversely, since our annotation system is solely based on P300 signatures which are task specific but target class agnostic. Therefore, it is not mandatory to train the EEGNet with object class-specific EEG responses. To validate this aspect, we trained and tested the EEGNet with EEG responses corresponding to different object categories. Table \ref{table:cresults} presents the F1 scores achieved for the five viewers with class-agnostic train and test EEG data. Note that only a marginal difference in annotation performance is noticeable with class-specific and class-agnostic EEG data across viewers. { Since we are using the pre-trained VGG-19 model exclusively to extract feature descriptors, it can be used without further fine tuning for any new target class categories.

\begin{table}[!h]
\begin{center}
\caption{Annotation performance with class-specific vs class-agnostic EEG data for five viewers.}
\label{table:cresults}
\begin{tabular}{|p{5cm}|p{1cm}|p{1cm}|p{1cm}|p{1cm}|p{1cm}|}
\hline
\textbf{F1 Score} & \textbf{ P1 } & \textbf{ P2 } & \textbf{ P3 } & \textbf{ P4 } & \textbf{ P5 } \\
\hline\hline
Class-specific train and test      & 0.88 & 0.86 & 0.89 & 0.87 & 0.88\\
Class-agnostic train and test	   & 0.85 & 0.85 & 0.84 & 0.86 & 0.86\\
\hline
\end{tabular}
\end{center}
\vspace{-.5cm}
\end{table}

%\subsection{Misclassified images}
%\label{misclassify}

%----------------------------------Conclusion-----------------------------------
\section{Conclusion}
 In order to facilitate large-scale image annotation efforts for computer vision and scene understanding applications, we propose an EEG-based fast image annotation system. Our annotation system exclusively relies on the P300 ERP signature, which is elicited upon the viewer detecting a pre-specified object class in the displayed image. A further outlier removal procedure based on binary feature-based clustering significantly improves annotation performance.
 
 Overall, our system achieves a peak F1-score of 0.88 with a 10 Hz annotation throughput. Another advantage of our method is that the P300 signature is specific to the target detection task, but not the underlying object class.Therefore, any novel image category can be annotated with existing models upon compiling the viewer EEG responses. Future work will focus on discovering and exploiting object-specific EEG signatures, and combining multiple human responses (\textit{e.g.}, EEG plus eye movements) for fine-grained object annotation and classification.

\bibliographystyle{IEEEtran}  
\bibliography{eegbib}

\end{document}